\documentclass{article}
\usepackage{ijcai26}

\usepackage{times}
\usepackage{soul}
\usepackage{url}
\usepackage[hidelinks]{hyperref}
\usepackage[utf8]{inputenc}
\usepackage[small]{caption}
\usepackage{graphicx}
\usepackage{amsmath}
\usepackage{amsthm}
\usepackage{booktabs}
\usepackage{algorithm}
\usepackage{algorithmic}
\usepackage[switch]{lineno}
\usepackage{amssymb} 
\usepackage{multirow}
\usepackage{makecell}
\usepackage{threeparttable}
\PassOptionsToPackage{table}{xcolor}
\usepackage[table]{xcolor}
\usepackage{array} 
\usepackage{adjustbox}
\usepackage{pifont}
\newcommand{\cmark}{\ding{51}} 
\newcommand{\xmark}{\ding{55}} 

\definecolor{myblue}{RGB}{236,243,250}   
\definecolor{mygray}{RGB}{242,242,242} 
\definecolor{myyellow}{RGB}{255,251,239} 
\definecolor{newyellow}{RGB}{255,245,217}
\definecolor{mygreen}{RGB}{238,247,233} 
\definecolor{mypink}{RGB}{255,247,251}
\definecolor{mypurple}{RGB}{247,247,255}
\definecolor{inkgreen}{RGB}{51,157,147}
\definecolor{mygreen2}{RGB}{243,251,250} 

\newcommand{\mname}[1]{\mbox{#1}}

\title{HyLoVQA: Dynamic Hypernetwork-Generated Low-Rank Adaptation for Continual Visual Question Answering}

\author{
    Yiran Wang$^1$
    \and
    Chenyi Xiong$^1$
    \and
    Ziyue Qin$^1$
    \and
    Miao Zhang$^1$
    \and
    Kui Xiao$^1$
    \and
    Zhifei Li$^{1,2,3}$\thanks{Corresponding author.}\\
    \affiliations
    $^1$School of Computer Science, Hubei University, Wuhan 430062, China\\
    $^2$Hubei Key Laboratory of Big Data Intelligent Analysis and Application (Hubei University), Wuhan 430062, China\\
    $^3$Key Laboratory of Intelligent Sensing System and Security (Hubei University), Ministry of Education, Wuhan 430062, China\\
    \emails
    \{yiranwang, xiongchenyi, qinziyue\}@stu.hubu.edu.cn,
    \{zhangmiao, xiaokui, zhifei1993\}@hubu.edu.cn
}

\begin{document}

\maketitle

\begin{abstract}
Continual Visual Question Answering (VQA) requires learning from non-stationary streams of visual inputs and questions while preserving past knowledge. Most prior methods adapt by updating a largely shared parameter set. This often leads to cross-level task interference, hindering accurate adaptation to the current task and object. To address this limitation, we propose HyLoVQA. It maintains a drift-resilient memory bank of anchors. The bank stores the content of visual objects and textual tasks, and they are updated using current input features. Conditioned on retrieved anchors, a hypernetwork generates lightweight Low-Rank Adaptation (LoRA) adapters. This ensures parameter efficiency, allowing the model to adapt to each task and object dynamically. Additionally, we formulate an alignment loss that aligns semantic discrepancies in the feature space with functional changes in the parameter space, thereby constraining LoRA adapters to remain focused on the current task and object. Extensive experiments on VQA v2 and NExT-QA under both standard and compositional settings demonstrate the superiority of HyLoVQA over prior state-of-the-art methods. The code is available at \url{https://github.com/HubuKG/HyLoVQA}.
\end{abstract}

\section{Introduction}
Visual Question Answering (VQA) is a central task in multimodal intelligence \cite{1,2}. 
Given an image (or video) and a natural-language question, a VQA model predicts the answer. 
Solving VQA requires grounding language in visual evidence and reasoning over the scene in a question-dependent way. 
This often calls for multi-step, compositional inference rather than direct recognition alone. 
Over time, the data distribution and question patterns may change, motivating robustness under shift and continual adaptation \cite{3,4,9}.

\begin{figure}[t]
\centering
\includegraphics[width=1.0\columnwidth]{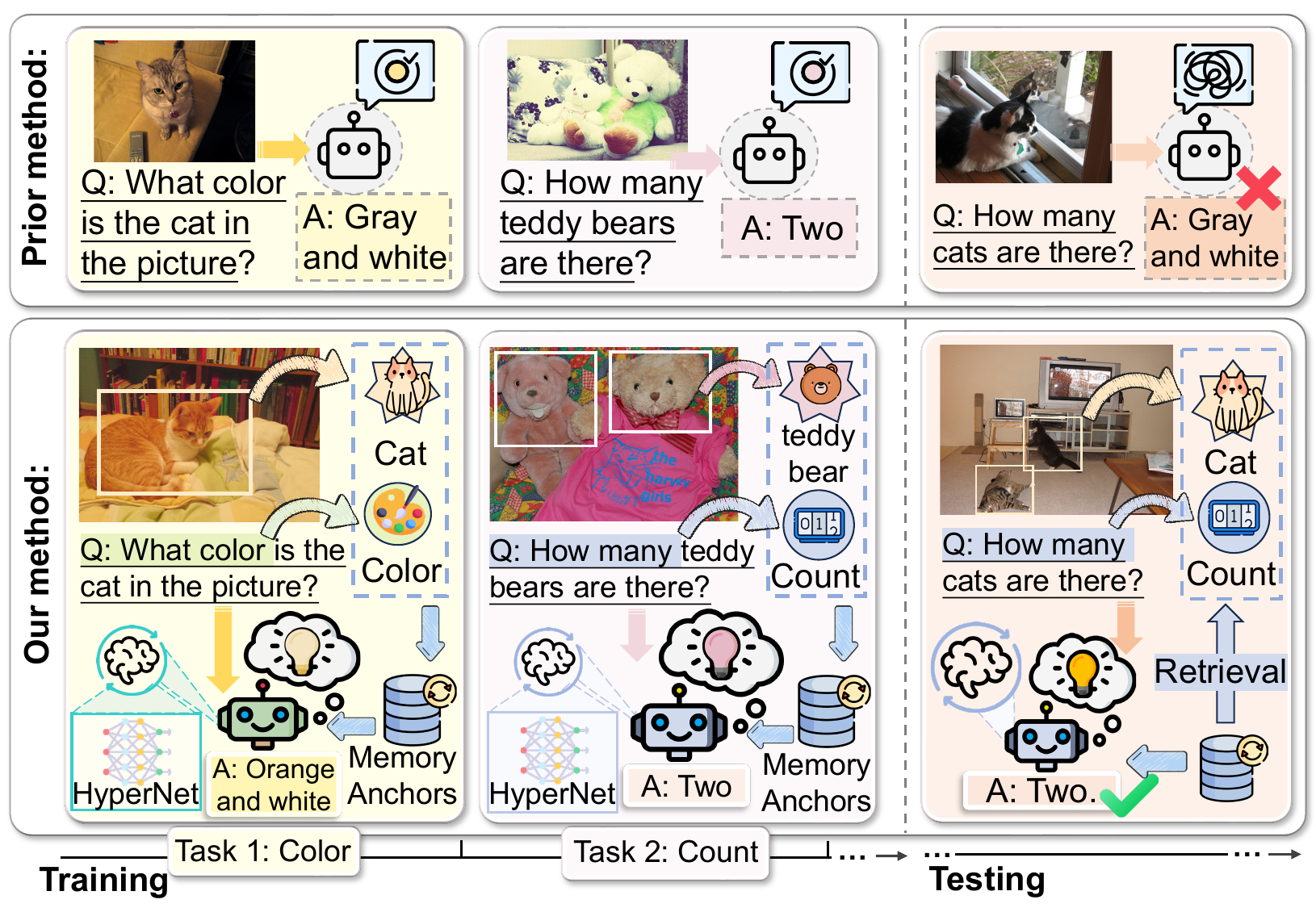} 
\caption{Motivation and overview. Top: Prior methods use shared-parameter continual updates, which induce task/object ambiguity (e.g., \textit{count} → \textit{color}). Bottom: HyLoVQA retrieves Memory Anchors and uses a HyperNet to generate LoRA adapters conditioned on the current task and object, reducing drift-induced interference.}
\label{figure:1}
\end{figure}

Continual learning (CL) offers a natural framework for continual VQA by learning from non-stationary data streams. It aims to acquire new knowledge while retaining prior capabilities. A central challenge in CL is the stability--plasticity trade-off: models must remain stable enough to avoid forgetting while staying plastic enough to rapidly integrate novel information. Classic CL methods include regularization, rehearsal, and architectural methods for unimodal learning. When applied to VQA, these ideas face added pressure from multimodal grounding and compositional reasoning. Shifts can arise not only in vision or language, but also in their cross-modal co-occurrences and required reasoning, which makes retention and updating more difficult than in unimodal classification. Recent continual VQA work adapts CL principles to multimodal grounding and compositional shifts; VQACL, for example, evaluates unseen skill--concept compositions~\cite{VQACL}.

Despite recent progress, most continual VQA methods still follow classic CL recipes. They adapt by updating the main model and mitigate forgetting with regularization or rehearsal~\cite{6,7}. Regularization improves stability but can reduce plasticity under large shifts. Rehearsal depends on the memory budget and how well stored samples remain representative as the stream evolves~\cite{42,44}.
These strategies can retain performance on past data. However, most existing methods adapt by updating a largely shared parameter set across the stream. This often induces cross-level task interference, where previously learned tasks disturb the current one. They also struggle to dynamically adapt to the current task and object. Here, the task is the capability required by the question (e.g., \textit{count}), and the object is the grounded entity in the input (e.g., \textit{a cat}). In this paper, a sample refers to a VQA instance, which can be viewed as a combination of a textual task and a visual object. As illustrated in the top of Figure~\ref{figure:1}, these methods can introduce task and object ambiguity, leading to incorrect answers.

To address this limitation, we propose HyLoVQA for continual VQA. It ensures parameter efficiency, enabling the model to adapt to each task and object dynamically while reducing drift-induced interference, as shown in Figure~\ref{figure:1}.
\textbf{First}, we maintain a Drift-Resilient Memory Anchor Bank. It stores compact anchors that capture the content of visual objects and textual tasks, and the anchors are updated using current input features to remain stable under non-stationary streams.
\textbf{Second}, we introduce a Hypernetwork-Generated LoRA Module, where a hypernetwork generates lightweight LoRA adapters from the retrieved anchors. This ensures parameter efficiency and allows the model to dynamically adapt to each task and object, while mitigating interference from shared backbone updates.
\textbf{Third}, we propose a Semantic--Functional Alignment loss that aligns semantic discrepancy (feature space) to functional change (parameter space). It avoids updates that deviate from the current task and object.

In this paper, we make the following contributions:
\begin{itemize}
  \item We develop a drift-resilient memory bank that stores anchor representations of visual objects and textual tasks, and updates them online using current input features.
  \item We generate lightweight LoRA adapters using a hypernetwork conditioned on retrieved anchors, enabling parameter-efficient and dynamic adaptation to each task and object.
  \item We formulate an alignment objective that couples feature-space semantic discrepancies with parameter-space functional changes, thereby constraining LoRA adapters to stay focused on the current task and object.
  \item We evaluate HyLoVQA on VQA v2 and NExT-QA under both standard and novel compositional test settings. On standard tests, HyLoVQA achieves 45.41\% AP / 2.82\% AF on VQA v2 and 42.43\% AP / 2.07\% AF on NExT-QA.
\end{itemize}

\section{Related Work}

\subsection{Visual Question Answering}
Visual Question Answering (VQA) asks a model to answer a natural-language question about an image\cite{21}. Early pipelines paired CNN image encoders with RNN question encoders and used attention for visual grounding\cite{56,33}. Fusion evolved from concatenation to bilinear pooling, with Multimodal Compact Bilinear pooling as an efficient high-order design\cite{24,23}. Object-centric representations further advanced VQA via region-level attention, where Bottom-Up and Top-Down attention over detected regions became a widely used baseline\cite{26,27}. Recently, Transformer-based architectures and large-scale vision-language pretraining have dominated VQA, improving multimodal representation learning and reasoning in the pretrain-then-finetune paradigm\cite{28}.

However, as VQA applications expand to more complex and dynamic scenarios, the assumptions of a fixed task and a stationary data distribution become fragile. For traditional VQA models, high scores on standard benchmarks may not transfer to VQA settings with distribution shifts~\cite{29,44}. Meanwhile, incomplete or noisy images and open-ended question formulations further amplify the limitations of fixed training setups. Therefore, VQA requires models that can adapt as tasks and data distributions keep changing, while avoiding forgetting previously learned knowledge. This has motivated continual learning for VQA to mitigate catastrophic forgetting and support long-term adaptation~\cite{30}.

\subsection{Continual Learning in VQA}
Continual Learning (CL) studies learning from sequential data streams whose distribution shifts over time~\cite{31,34}. A central challenge is the stability--plasticity trade-off~\cite{46}. Models must acquire new knowledge while retaining old knowledge. CL has been widely explored in vision and language, and recent efforts extend it to multimodal settings. To mitigate catastrophic forgetting, most methods fall into three categories: regularization, rehearsal, and architectural approaches. Regularization methods constrain updates to parameters that are important for past tasks, such as MAS and NPC~\cite{MAS,NPC}. Rehearsal methods replay stored samples to stabilize training, such as ER and DER~\cite{ER,DER,VS}. Architectural methods are commonly studied in unimodal continual learning\cite{54}.

Recent continual VQA methods tackle evolving visual concepts and language. VQACL proposes a dual-level task stream and compositional testing. QUAD and ProtoGroup reduce memory via attention consistency and multi-prototype grouping. PROOF mitigates interference with expandable projections, while CLT-VQA targets long-tailed continual VQA via prototype balancing and feature alignment\cite{QUAD,ProtoGroup,PROOF,CLT-VQA}. Yet these approaches typically keep a fixed backbone, so interference from shared parameters can still accumulate over time. In contrast, our work reduces cross-level task interference and improves accurate adaptation to the current task and object. By preserving past knowledge while maintaining parameter efficiency, it provides a more reliable and adaptive solution for long-term continual VQA.

\begin{figure*}[t]
\centering
\includegraphics[width=1.0\linewidth]{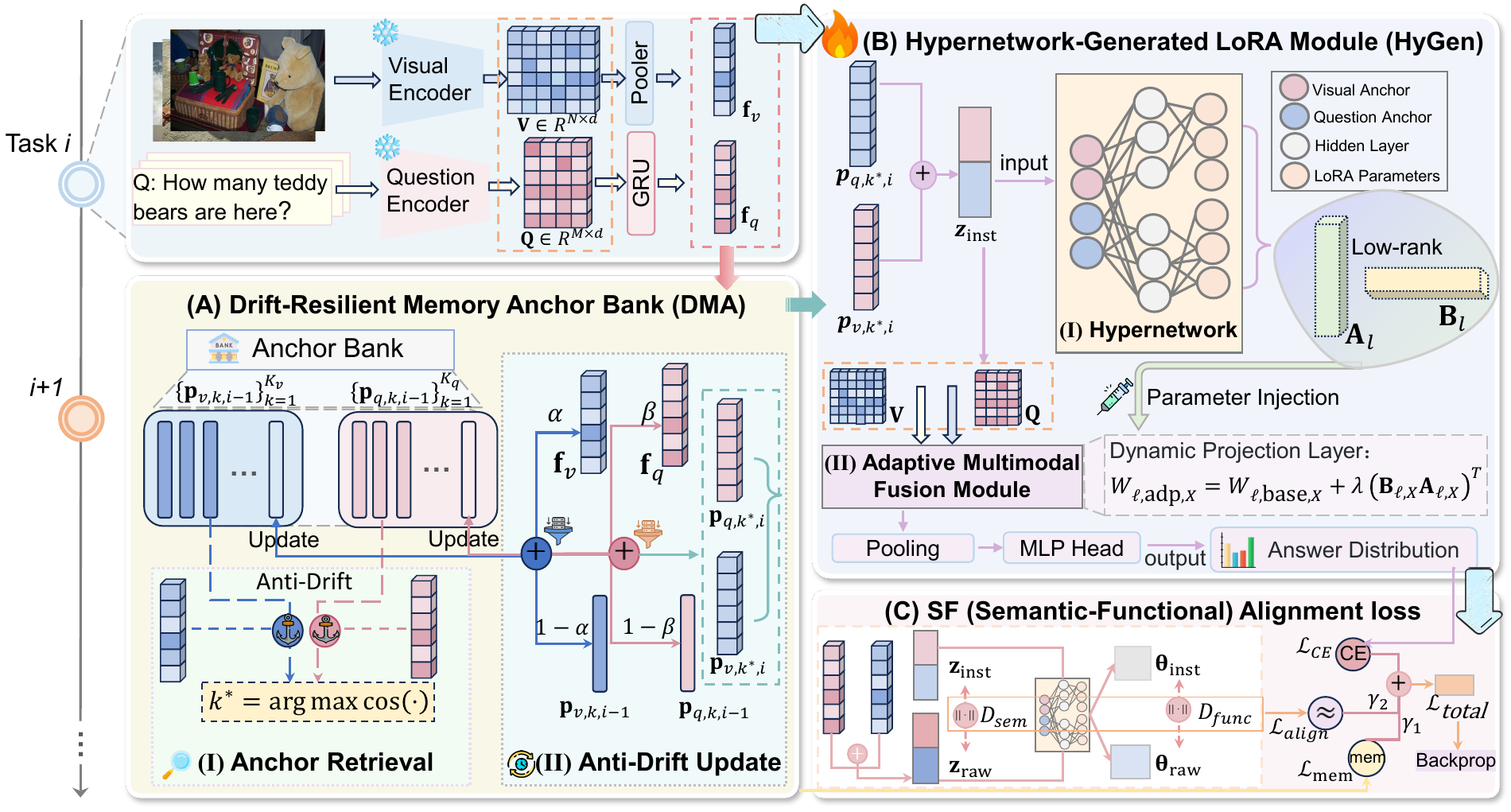} 
\caption{Overview of HyLoVQA. (A) Drift-Resilient Memory Anchor Bank stores compact anchors for visual objects and textual tasks and updates them with current input features for stability. (B) Hypernetwork-Generated LoRA Module generates anchor-conditioned lightweight LoRA adapters for parameter-efficient, task/object-specific adaptation with reduced interference on the shared backbone. (C) Semantic--Functional alignment loss constrains training by aligning semantic discrepancy (feature space) with functional change (parameter space), avoiding LoRA adapters that deviate from the current sample.}
\label{fig:overview}
\end{figure*}

\section{Methodology}
\label{sec:method}
We propose HyLoVQA, a lightweight continual learning framework (Figure~\ref{fig:overview}) for visual question answering, addressing dynamic adaptation, memory robustness under drift, and stable knowledge updates. By integrating a Drift-Resilient Memory Anchor Bank, a Hypernetwork-Generated LoRA Module, and a Semantic--Functional alignment loss, HyLoVQA enables parameter-efficient, task/object-specific adaptation while reducing interference on the shared backbone. Module details are described below.

\subsection{Drift-Resilient Memory Anchor Bank}
\label{sec:DMA}

The Drift-Resilient Memory Anchor Bank (DMA) stores compact, modality-specific anchors that summarize visual objects and textual tasks. For each incoming sample, DMA retrieves the most relevant visual and question anchors to form an anchor-conditioned context, which then conditions the hypernetwork to generate LoRA adapters. This yields an explicit long-term memory for continual VQA.

Multimodal representations are constructed from the input image and question as follows. Given an image $I$, $N$ salient regions are extracted and each region is embedded into a $d$-dimensional token by linearly projecting its RoI feature $\mathbf{f}_{j,\mathrm{roi}}\in\mathbb{R}^{d_{\mathrm{roi}}}$, adding a projected normalized bounding box $\mathbf{b}_{j}=[x_1,y_1,x_2,y_2]\in\mathbb{R}^{4}$ (each coordinate in $[0,1]$), and a learnable region-id embedding $\mathbf{e}_{j,\mathrm{id}}\in\mathbb{R}^{d}$. The resulting region token is denoted by $\mathbf{v}_j\in\mathbb{R}^d$. Stacking all region tokens yields $\mathbf{V}\in\mathbb{R}^{N\times d}$, and a global visual summary is computed by mean pooling:
\begin{equation}
\mathbf{f}_{v}=\frac{1}{N}\sum_{j=1}^{N}\mathbf{v}_{j}.
\label{eq:vis_global}
\end{equation}

For a question $Q$ with $M$ tokens, a text encoder outputs token features $\mathbf{Q}\in\mathbb{R}^{M\times d}$, and a global question summary is obtained as
\begin{equation}
\mathbf{f}_{q}=\mathrm{Pool}(\mathbf{Q}),
\label{eq:txt_global}
\end{equation}
where $\mathrm{Pool}(\cdot)$ denotes a pooling operator.

Within task $i$, a sample denotes an image--question pair. DMA updates the anchor bank online per sample, and the within-task index is omitted by using $\mathcal{P}_{v,i}$ and $\mathcal{P}_{q,i}$ to denote the current bank. At the beginning of task $i$, anchors are inherited from the previous task (for $i{=}1$, $\mathcal{P}_{v,0}=\{\mathbf{f}_v^{(k)}\}_{k=1}^{K_v}$ and $\mathcal{P}_{q,0}=\{\mathbf{f}_q^{(k)}\}_{k=1}^{K_q}$):
\begin{equation}
\left\{
\begin{aligned}
\mathcal{P}_{v,i-1} &= \{\mathbf{p}_{v,k,i-1}\}_{k=1}^{K_{v}}\\
\mathcal{P}_{q,i-1} &= \{\mathbf{p}_{q,k,i-1}\}_{k=1}^{K_{q}}
\end{aligned}
\right..
\label{eq:anchor_bank}
\end{equation}

For each incoming sample, the most relevant anchors are retrieved by cosine similarity
$\mathrm{cos}(\mathbf{a},\mathbf{b})=\frac{\mathbf{a}^{\top}\mathbf{b}}{\|\mathbf{a}\|_{2}\,\|\mathbf{b}\|_{2}}$:
\begin{equation}
\left\{
\begin{aligned}
k_{v}^{\ast}
&=
\arg\max_{k\in\{1,\dots,K_{v}\}}
\mathrm{cos}\!\left(\mathbf{f}_{v},\mathbf{p}_{v,k,i-1}\right)\\
k_{q}^{\ast}
&=
\arg\max_{k\in\{1,\dots,K_{q}\}}
\mathrm{cos}\!\left(\mathbf{f}_{q},\mathbf{p}_{q,k,i-1}\right)
\end{aligned}
\right..
\label{eq:anchor_retrieval}
\end{equation}

Only the retrieved anchors are updated using momentum:
\begin{equation}
\left\{
\begin{aligned}
\mathbf{p}_{v,k_{v}^{\ast},i}
&\leftarrow
\alpha\,\mathbf{p}_{v,k_{v}^{\ast},i-1} + (1-\alpha)\,\mathbf{f}_{v}\\
\mathbf{p}_{q,k_{q}^{\ast},i}
&\leftarrow
\beta\,\mathbf{p}_{q,k_{q}^{\ast},i-1} + (1-\beta)\,\mathbf{f}_{q}
\end{aligned}
\right.,
\label{eq:anchor_update}
\end{equation}
where $\alpha,\beta\in[0,1)$ are sample-adaptive coefficients. For all non-retrieved anchors, they are kept unchanged, e.g., $\mathbf{p}_{v,k,i}=\mathbf{p}_{v,k,i-1}$ for $k\neq k_v^\ast$ and $\mathbf{p}_{q,k,i}=\mathbf{p}_{q,k,i-1}$ for $k\neq k_q^\ast$, yielding the updated bank $\mathcal{P}_{v,i}$ and $\mathcal{P}_{q,i}$.

Finally, the instance context is defined as
\begin{equation}
\mathbf{z}_{\mathrm{inst},i}
=
\left[
\mathbf{p}_{v,k_{v}^{\ast},i}\,;\,
\mathbf{p}_{q,k_{q}^{\ast},i}
\right].
\label{eq:z_inst}
\end{equation}

To further mitigate representation drift in continual learning, we introduce a
memory consistency loss. We use $\mathcal{L}_{\mathrm{mem}}$ to denote the aggregate of
modality-specific memory terms:
\begin{equation}
\mathcal{L}_{\mathrm{mem}}
=
\left(1-\mathrm{cos}\!\left(\mathbf{f}_{v},\mathbf{p}_{v,k_{v}^{\ast},i}\right)\right)
+
\left(1-\mathrm{cos}\!\left(\mathbf{f}_{q},\mathbf{p}_{q,k_{q}^{\ast},i}\right)\right).
\label{eq:loss_mem}
\end{equation}

In the next section~\ref{sec:hypergen}, it is detailed how $\mathbf{z}_{\mathrm{inst},i}$ conditions the hypernetwork to generate LoRA adapters.

\subsection{Hypernetwork-Generated LoRA Module}
\label{sec:hypergen}
\label{sec:fusion}

The Hypernetwork-Generated LoRA Module (HyGen) enables continual VQA adaptation by generating lightweight LoRA adapters for a largely shared backbone. A hypernetwork produces these updates conditioned on the anchors retrieved from DMA, allowing sample-wise, dynamic adaptation under non-stationary streams. This design adapts the model by injecting Hypernetwork-Generated LoRA adapters into a shared backbone, maintaining parameter efficiency while reducing interference across tasks.

The construction starts from generating LoRA factors for each incoming sample within task $i$, conditioned on its instance representation $\mathbf{z}_{\mathrm{inst},i}$. Let $H_{\mathbf{\phi}}(\cdot)$ denote a hypernetwork with parameters $\mathbf{\phi}$.
To enable layer-specific generation, we additionally introduce a learned layer embedding $\mathbf{e}_{\ell}\in\mathbb{R}^{d_{\ell}}$.
For each fusion layer $\ell\in\{1,\dots,L\}$, we generate LoRA factors for cross-attention projections as
\begin{equation}
\left\{
\mathbf{A}_{\ell,X},
\mathbf{B}_{\ell,X}
\right\}_{X\in\{Q,K,V\}}
=
H_{\mathbf{\phi}}\!\left(\left[\mathbf{z}_{\mathrm{inst},i};\mathbf{e}_{\ell}\right]\right),
\label{eq:hyper_out}
\end{equation}
where $X\in\{Q,K,V\}$ denotes the projection type (query/key/value), $\mathbf{A}_{\ell,X}\in\mathbb{R}^{r\times d_{\mathrm{in}}}$, $\mathbf{B}_{\ell,X}\in\mathbb{R}^{d_{\mathrm{out}}\times r}$, and the rank satisfies $r\ll \min(d_{\mathrm{in}},d_{\mathrm{out}})$ (typically $d_{\mathrm{in}}=d_{\mathrm{out}}=d$).

Given the generated factors, the frozen cross-attention projections are augmented to obtain sample-adaptive projections.
We represent each sequence as a matrix whose rows are token embeddings, and apply linear projections by right-multiplication.
Let $\mathbf{W}_{\ell,\mathrm{base},X}\in\mathbb{R}^{d_{\mathrm{in}}\times d_{\mathrm{out}}}$ denote the frozen base projection.
The sample-adaptive projection is formed as
\begin{equation}
\mathbf{W}_{\ell,\mathrm{adp},X}
=
\mathbf{W}_{\ell,\mathrm{base},X}
+
\lambda\,\left(\mathbf{B}_{\ell,X}\mathbf{A}_{\ell,X}\right)^{\top},
\label{eq:lora_inject}
\end{equation}
where $\lambda\ge 0$ controls the adaptation strength.

In parallel to the parameter adaptation, the retrieved anchors are injected as memory-guided global context tokens by prepending them to the original sequences (concatenation along the sequence-length dimension).
Here $\mathbf{Q}\in\mathbb{R}^{n_q\times d}$ and $\mathbf{V}\in\mathbb{R}^{n_v\times d}$ denote the input token sequences on the query side and value side of cross-attention, respectively (note that $\mathbf{V}$ here denotes a sequence, not the value projection).
We compute
\begin{equation}
\left\{
\begin{aligned}
\mathbf{g}_{v} &= \mathbf{W}_{g}\mathbf{p}_{v,k_{v}^{\ast},i} \\
\hat{\mathbf{V}} &= \left[\mathbf{g}_{v};\mathbf{V}\right]
\end{aligned}
\right.,
\qquad
\left\{
\begin{aligned}
\mathbf{g}_{q} &= \mathbf{W}_{g}\mathbf{p}_{q,k_{q}^{\ast},i} \\
\hat{\mathbf{Q}} &= \left[\mathbf{g}_{q};\mathbf{Q}\right]
\end{aligned}
\right.,
\label{eq:ctx_tokens}
\end{equation}
where $\mathbf{p}_{v,k_{v}^{\ast},i}\in\mathbb{R}^{d}$ and $\mathbf{p}_{q,k_{q}^{\ast},i}\in\mathbb{R}^{d}$ are the retrieved anchor embeddings for task $i$,
and $\mathbf{W}_{g}\in\mathbb{R}^{d\times d}$ is learnable.
Accordingly, $\mathbf{g}_{v},\mathbf{g}_{q}\in\mathbb{R}^{d}$ are global context tokens.

With the augmented sequences and sample-adaptive projections in place, the model performs adaptive multimodal fusion and outputs the answer distribution.
At layer $\ell$, we compute cross-attention with adaptive projections as
\begin{equation}
\left\{
\begin{aligned}
\mathbf{S}_{\ell}
&=
\frac{
\left(\hat{\mathbf{Q}}\mathbf{W}_{\ell,\mathrm{adp},Q}\right)
\left(\hat{\mathbf{V}}\mathbf{W}_{\ell,\mathrm{adp},K}\right)^{\top}
}{
\sqrt{d_{k}}
}\\
\mathbf{H}_{\ell}
&=
\mathrm{Softmax}(\mathbf{S}_{\ell})
\left(\hat{\mathbf{V}}\mathbf{W}_{\ell,\mathrm{adp},V}\right)
\end{aligned}
\right.,
\label{eq:cross_attn}
\end{equation}
where $d_k$ is the key dimension used for scaling (typically $d_k=d/h$ for $h$ attention heads), and $\mathrm{Softmax}(\cdot)$ is applied row-wise.
For brevity, we omit the standard multi-head reshape and per-head projections in the notation.

After $L$ fusion layers, we pool the fused sequence into $\mathbf{e}_{\mathrm{out}}$ and
predict an answer distribution $\mathbf{\pi}\in[0,1]^C$ via an MLP with Softmax.
We construct a soft target distribution $\mathbf{t}\in[0,1]^C$ from multiple human
annotations.
\begin{equation}
\mathcal{L}_{\mathrm{CE}}
=
-\sum_{c=1}^{C} t_c \log \mathbf{\pi}_c.
\label{eq:l_ce}
\end{equation}

We optimize the model by minimizing $\mathcal{L}_{\mathrm{CE}}$.

\begin{table*}[t]
  \centering
  \small
  \renewcommand{\arraystretch}{0.85}
  \setlength{\tabcolsep}{5.5pt}

  \begin{adjustbox}{width=\textwidth}
    \begin{tabular}{c|ccccc|ccccc}
      \toprule
      \multirow{3}{*}{\textbf{Methods}} &
      \multicolumn{5}{c|}{\cellcolor{myyellow}\textbf{VQA v2}} &
      \multicolumn{5}{c}{\cellcolor{mygreen}\textbf{NExT-QA}} \\
      \cmidrule(lr){2-6}\cmidrule(lr){7-11}

      & \multicolumn{2}{c}{\textbf{Standard Test}} &
        \multicolumn{2}{c}{\textbf{Novel Comp. Test}} &
        \multirow{2}{*}{\textbf{\#Mem}} &
        \multicolumn{2}{c}{\textbf{Standard Test}} &
        \multicolumn{2}{c}{\textbf{Novel Comp. Test}} &
        \multirow{2}{*}{\textbf{\#Mem}} \\
      \cmidrule(lr){2-3}\cmidrule(lr){4-5}\cmidrule(lr){7-8}\cmidrule(lr){9-10}

      & \textbf{AP($\uparrow$)} & \textbf{AF($\downarrow$)} &
        \textbf{AP($\uparrow$)} & \textbf{AF($\downarrow$)} &
        &
        \textbf{AP($\uparrow$)} & \textbf{AF($\downarrow$)} &
        \textbf{AP($\uparrow$)} & \textbf{AF($\downarrow$)} & \\
      \midrule

      \mname{Vanilla}                &
      14.49 & 30.80 & 11.79 & 27.16 & 5000 &
      11.97 & 26.14 & 12.59 & 28.04 & 500 \\

      \mname{EWC (PNAS'17)}          &
      15.77 & 30.62 & 12.83 & 28.16 & 5000 &
      13.01 & 24.06 & 11.91 & 27.44 & 500 \\

      \mname{MAS (ECCV'18)}          &
      20.56 & 11.16 & 23.90 &  6.24 & 5000 &
      18.04 & 10.07 & 21.12 & 10.09 & 500 \\

      \mname{ER (MTLR'19)}           &
      36.99 &  5.99 & 33.78 &  5.76 & 5000 &
      30.55 &  4.91 & 32.20 &  5.57 & 500 \\

      \mname{DER (NeurIPS'20)}       &
      35.35 &  8.62 & 31.52 &  8.59 & 5000 &
      26.17 &  5.12 & 21.56 & 12.68 & 500 \\

      \mname{VS (CVPR'22)}           &
      34.03 &  8.79 & 32.96 &  5.78 & 5000 &
      28.13 &  4.45 & 29.47 &  6.14 & 500 \\

      \mname{VQACL (CVPR'23)}        &
      38.77 &  3.96 & 35.40 &  4.90 & 5000 &
      32.27 &  3.00 & 34.22 &  3.80 & 500 \\

      \mname{QUAD (ICCV'25)}         &
      39.25 &  4.91 & \cellcolor{mygray}\underline{40.00} & 3.81 & 5000 &
      31.70 &  2.91 & 33.21 &  4.16 & 500 \\

      \mname{ProtoGroup (ICASSP'25)} &
      39.81 & 2.87 & 36.90 & 4.06 & 5000 &
      35.29 & 2.15 &
      35.79 & \cellcolor{mygray}\underline{3.17} & 500 \\

      \mname{CLT-VQA (ICCV’25)}      &
      \cellcolor{mygray}\underline{40.69} & \cellcolor{mygray}\underline{2.11} &
      38.46 & \cellcolor{mygray}\underline{3.51} & 5000 &
      \cellcolor{mygray}\underline{35.77} &  \cellcolor{mygray}\underline{2.09} & \cellcolor{mygray}\underline{35.86} &  3.64 & 500 \\

      \noalign{\vskip 0.6ex \hrule height 0.4pt \vskip 0.4ex}

      \rowcolor{myblue}
      \textbf{HyLoVQA (Ours)} &
      \textbf{45.41} & \textbf{2.82} & \textbf{44.96} & \textbf{2.62} & \textbf{5000} &
      \textbf{42.43} & \textbf{2.07} & \textbf{43.05} & \textbf{3.10} & \textbf{500} \\
      \bottomrule
    \end{tabular}
  \end{adjustbox}

  \caption{Comparison on VQA v2 and NExT-QA under Standard and Novel Composition tests.}
  \label{tab:main_results}
\end{table*}

\begin{table*}[t]
  \centering
  \small
  \renewcommand{\arraystretch}{0.75}
  \setlength{\tabcolsep}{4.2pt} 

  \begin{adjustbox}{width=\textwidth}
    \begin{tabular}{c|c|cc|cc|cc|cc|cc|cc}
      \toprule
      \multirow{2}{*}{\textbf{Dataset}} & \multirow{2}{*}{\textbf{Methods}} &
      \multicolumn{2}{c|}{\cellcolor{mypink}\textbf{Group-1}} &
      \multicolumn{2}{c|}{\cellcolor{mypink}\textbf{Group-2}} &
      \multicolumn{2}{c|}{\cellcolor{mypink}\textbf{Group-3}} &
      \multicolumn{2}{c|}{\cellcolor{mypink}\textbf{Group-4}} &
      \multicolumn{2}{c|}{\cellcolor{mypink}\textbf{Group-5}} &
      \multicolumn{2}{c}{\cellcolor{mypurple}\textbf{Avg}} \\
      \cmidrule(lr){3-4}\cmidrule(lr){5-6}\cmidrule(lr){7-8}\cmidrule(lr){9-10}\cmidrule(lr){11-12}\cmidrule(lr){13-14}
      & &
      \textbf{Novel} & \textbf{Seen} &
      \textbf{Novel} & \textbf{Seen} &
      \textbf{Novel} & \textbf{Novel} &
      \textbf{Seen}  & \textbf{Seen}  &
      \textbf{Novel} & \textbf{Seen} &
      \textbf{Novel} & \textbf{Seen} \\
      \midrule

      \multirow{8}{*}{\shortstack{\textbf{VQA}\\\textbf{v2}}} &
      \mname{ER}  & 34.52 & 37.03 & 33.40 & 35.55 & 34.79 & 34.20 & 33.86 & 35.02 & 32.34 & 35.91 & 33.78 & 35.54 \\
      & \mname{DER} & 30.80 & 29.89 & 32.19 & 33.24 & 34.88 & 34.08 & 29.60 & 30.90 & 30.14 & 32.56 & 31.52 & 32.13 \\
      & \mname{VS}  & 33.35 & 33.87 & 33.18 & 32.21 & 34.50 & 33.84 & 31.29 & 33.98 & 32.46 & 33.87 & 32.96 & 33.55 \\
      & \mname{VQACL} & 36.12 & 37.99 & 35.39 & 36.92 & 36.26 & 35.16 & 34.85 & 35.64 & 34.36 & 36.28 & 35.40 & 36.40 \\
      & \mname{QUAD} &
      \cellcolor{mygray}\underline{39.19} & \cellcolor{mygray}\underline{41.06} &
      38.40 & \cellcolor{mygray}\underline{39.50} &
      \cellcolor{mygray}\underline{43.15} & 39.19 &
      \cellcolor{mygray}\underline{40.01} & \cellcolor{mygray}\underline{40.72} &
      \cellcolor{mygray}\underline{39.20} & \cellcolor{mygray}\underline{40.62} &
      \cellcolor{mygray}\underline{40.00} & \cellcolor{mygray}\underline{40.21} \\

      & \mname{ProtoGroup} & 36.23 & 36.85 & 35.76 & 37.91 & 38.70 &
      \cellcolor{mygray}\underline{39.57} &
      36.74 & 37.85 & 37.65 & 38.19 & 36.90 & 37.78 \\

      & \mname{CLT-VQA} & 37.78 & 37.91 & \cellcolor{mygray}\underline{38.95} & 38.11 & 36.55 & 38.68 & 39.72 & 39.94 & 36.88 & 38.78 & 38.46 & 39.51 \\




      \addlinespace[0.5ex]

      & \textbf{HyLoVQA} &
      \textbf{40.15} & \textbf{41.90} &
      \textbf{41.05} & \textbf{41.56} &
      \textbf{45.49} & \textbf{44.90} &
      \textbf{45.55} & \textbf{45.33} &
      \textbf{44.98} & \textbf{45.26} &
      \textbf{44.96} & \textbf{45.30} \\

      \specialrule{0.4pt}{0.2ex}{0.2ex}

      & \cellcolor{myblue}{+$\Delta$} &
      \cellcolor{myblue}{0.96} & \cellcolor{myblue}{0.84} &
      \cellcolor{myblue}{2.10} & \cellcolor{myblue}{2.06} &
      \cellcolor{myblue}{2.34} & \cellcolor{myblue}{5.33} &
      \cellcolor{myblue}{5.54} & \cellcolor{myblue}{4.61} &
      \cellcolor{myblue}{5.78} & \cellcolor{myblue}{4.64} &
      \cellcolor{myblue}{4.96} & \cellcolor{myblue}{5.09} \\

      \midrule

      \multirow{7}{*}{\shortstack{\textbf{NExT}\\\textbf{-QA}}} &
      \mname{ER}  & 31.86 & 34.51 & 32.36 & 35.08 & 29.50 & 34.30 & 33.57 & 33.30 & 33.71 & 32.91 & 32.20 & 34.02 \\
      & \mname{DER} & 27.56 & 26.09 & 26.14 & 24.54 & 23.53 & 26.43 &  9.30 &  9.79 & 21.26 & 23.74 & 21.56 & 21.38 \\
      & \mname{VS}  & 31.42 & 30.88 & 29.17 & 31.26 & 25.23 & 26.10 & 30.01 & 29.10 & 31.54 & 31.79 & 29.47 & 29.83 \\
      & \mname{VQACL} & 35.50 & 35.54 & 33.97 & 35.91 & 31.34 & 35.62 & 34.08 & 33.57 & 36.71 & 33.46 & 34.22 & 34.82 \\
      & \mname{QUAD} & 33.42 & 33.92 & 32.02 & 35.42 & 31.78 & 36.36 & 32.98 & 33.34 &
      \cellcolor{mygray}\underline{37.84} &
      34.06 & 33.21 & 34.62 \\

      & \mname{ProtoGroup} & 35.79 & 36.18 & 34.56 & 36.83 & 32.90 & 36.58 & 35.21 & 35.60 & 36.92 & 34.40 & 35.79 & 36.48 \\
      & \mname{CLT-VQA} &
      \cellcolor{mygray}\underline{36.14} & \cellcolor{mygray}\underline{36.46} &
      \cellcolor{mygray}\underline{35.98} & \cellcolor{mygray}\underline{37.03} &
      \cellcolor{mygray}\underline{38.71} & \cellcolor{mygray}\underline{37.32} &
      \cellcolor{mygray}\underline{36.44} & \cellcolor{mygray}\underline{36.92} &
      36.39 & \cellcolor{mygray}\underline{35.89} &
      \cellcolor{mygray}\underline{35.86} & \cellcolor{mygray}\underline{36.79} \\

      \addlinespace[0.5ex]

      & \textbf{HyLoVQA} &
      \textbf{40.23} & \textbf{41.56} &
      \textbf{41.77} & \textbf{42.21} &
      \textbf{42.97} & \textbf{43.28} &
      \textbf{42.39} & \textbf{43.96} &
      \textbf{43.15} & \textbf{43.08} &
      \textbf{43.05} & \textbf{43.97} \\

      \specialrule{0.4pt}{0.2ex}{0.2ex}

      & \cellcolor{myblue}{+$\Delta$} &
      \cellcolor{myblue}{4.09} & \cellcolor{myblue}{5.10} &
      \cellcolor{myblue}{5.79} & \cellcolor{myblue}{5.18} &
      \cellcolor{myblue}{+4.26} & \cellcolor{myblue}{+5.96} &
      \cellcolor{myblue}{5.95} & \cellcolor{myblue}{7.04} &
      \cellcolor{myblue}{+5.31} & \cellcolor{myblue}{7.19} &
      \cellcolor{myblue}{+7.19} & \cellcolor{myblue}{7.18} \\

      \bottomrule
    \end{tabular}
  \end{adjustbox}

  \caption{Fine-grained VQA performance AP (\%) on the Novel and Seen skill-concept compositions of VQA v2 and NExT-QA.}
  \label{tab:group_results}
\end{table*}

\subsection{Semantic--Functional Alignment loss}
\label{sec:align}
\label{sec:obj}

The Semantic--Functional (SF) Alignment loss stabilizes anchor-conditioned adaptation by enforcing consistency between semantic and functional changes. Specifically, it matches the semantic distance between the current input features and the retrieved anchors with the functional change measured in parameter space between the corresponding Hypernetwork-Generated LoRA adapters. The loss discourages inconsistent parameter updates and improves long-term robustness.

For task $i$, $\mathbf{z}_{\mathrm{raw},i}$ is constructed by concatenating $\mathbf{f}_v$ and $\mathbf{f}_q$. The semantic discrepancy between the anchor-conditioned context and $\mathbf{z}_{\mathrm{raw},i}$ is measured using cosine distance:
\begin{equation}
D_{\mathrm{sem}}
\left(
\mathbf{z}_{\mathrm{inst},i},\mathbf{z}_{\mathrm{raw},i}
\right)
=
1-\mathrm{cos}\!\left(\mathbf{z}_{\mathrm{inst},i},\mathbf{z}_{\mathrm{raw},i}\right).
\label{eq:d_sem}
\end{equation}

To measure functional discrepancy, the hypernetwork outputs across all layers are vectorized, where $\mathrm{vec}(\cdot)$ concatenates all generated matrices into a single vector:
\begin{equation}
\left\{
\begin{aligned}
\mathbf{\theta}_{\mathrm{inst}}
&=
\mathrm{vec}\!\left(H_{\mathbf{\phi}}(\mathbf{z}_{\mathrm{inst},i})\right)\\
\mathbf{\theta}_{\mathrm{raw}}
&=
\mathrm{vec}\!\left(H_{\mathbf{\phi}}(\mathbf{z}_{\mathrm{raw},i})\right)
\end{aligned}
\right.,
\label{eq:theta_vec}
\end{equation}
where $m$ is the total number of generated scalar parameters after vectorization, and a normalized $\ell_{2}$ distance in parameter space is defined as:
\begin{equation}
D_{\mathrm{func}}
\left(
\mathbf{\theta}_{\mathrm{inst}},\mathbf{\theta}_{\mathrm{raw}}
\right)
=
\frac{\left\|\mathbf{\theta}_{\mathrm{inst}}-\mathbf{\theta}_{\mathrm{raw}}\right\|_{2}}{\sqrt{m}}.
\label{eq:d_func}
\end{equation}

Dual-Space Consistency is used as an auxiliary objective: $\mathcal{L}_{\mathrm{align}}$ encourages agreement between the anchor-conditioned semantic shift ($\mathbf{z}_{\mathrm{inst}}$ vs.\ $\mathbf{z}_{\mathrm{raw}}$) and the induced change in parameter space:
\begin{equation}
\mathcal{L}_{\mathrm{align}}
=
\left(
D_{\mathrm{sem}}
\left(
\mathbf{z}_{\mathrm{inst},i},\mathbf{z}_{\mathrm{raw},i}
\right)
-
D_{\mathrm{func}}
\left(
\mathbf{\theta}_{\mathrm{inst}},\mathbf{\theta}_{\mathrm{raw}}
\right)
\right)^{2}.
\label{eq:l_align}
\end{equation}
The overall objective is
\begin{equation}
\mathcal{L}
=
\mathcal{L}_{\mathrm{CE}}
+
\gamma_{\mathrm{1}}\,\mathcal{L}_{\mathrm{mem}}
+
\gamma_{\mathrm{2}}\,\mathcal{L}_{\mathrm{align}},
\label{eq:total_loss}
\end{equation}
where $\gamma_{1}$ and $\gamma_{2}$ control the relative weights of
$\mathcal{L}_{\mathrm{mem}}$ and $\mathcal{L}_{\mathrm{align}}$, respectively. We minimize $\mathcal{L}_{\mathrm{total}}$ by backpropagation, updating all trainable modules while keeping the base backbone projections frozen.

\section{Experiments}
\label{sec:exp}

To comprehensively evaluate HyLoVQA in continual VQA, we conduct experiments addressing six research questions:
\begin{itemize}
    \item \textbf{RQ1:} How does HyLoVQA compare to baselines on Standard and Novel Composition tests, overall and fine-grained?
    \item \textbf{RQ2:} What is the contribution of each component to overall effectiveness and forgetting?
    \item \textbf{RQ3:} How effective are HyLoVQA's sample-specific LoRA adapters, and what evidence supports this?
    \item \textbf{RQ4:} What is the impact of DMA capacity on retention and forgetting across tasks?
    \item \textbf{RQ5:} What is HyLoVQA's sensitivity to key hyperparameters such as anchor momentum?
    \item \textbf{RQ6:} What failure modes of prior methods are revealed by case studies, and how does HyLoVQA address them?
\end{itemize}

\begin{table}[!t]
\centering
\small
\renewcommand{\arraystretch}{0.92}
\setlength{\tabcolsep}{4pt}

\begin{adjustbox}{width=\columnwidth}
\begin{tabular}{c|c|c|c c|c c}
\toprule
\multicolumn{3}{c|}{\textbf{Method}} &
\multicolumn{2}{c|}{\textbf{\cellcolor{myyellow}VQA v2}} &
\multicolumn{2}{c}{\textbf{\cellcolor{mygreen}NExT-QA}} \\
\cmidrule(r){1-3}\cmidrule(lr){4-5}\cmidrule(l){6-7}
\textbf{DMA} & \textbf{HyGen} & \textbf{Align.} &
\textbf{AP} ($\uparrow$) & \textbf{AF} ($\downarrow$) &
\textbf{AP} ($\uparrow$) & \textbf{AF} ($\downarrow$) \\
\midrule
\cmark & \xmark & \xmark &
41.77\textsubscript{\textcolor{inkgreen}{(-3.64)}} & 3.46\textsubscript{\textcolor{inkgreen}{(+0.64)}} & 39.14\textsubscript{\textcolor{inkgreen}{(-3.29)}} & 3.79\textsubscript{\textcolor{inkgreen}{(+1.72)}} \\
\cmark & \cmark & \xmark & 44.96\textsubscript{\textcolor{inkgreen}{(-0.45)}} & 2.95\textsubscript{\textcolor{inkgreen}{(+0.13)}} & 41.65\textsubscript{\textcolor{inkgreen}{(-0.78)}} & 2.52\textsubscript{\textcolor{inkgreen}{(+0.45)}} \\

\addlinespace[0.5ex]

\rowcolor{myblue}
\cmark & \cmark & \cmark & \textbf{45.41} & \textbf{2.82} & \textbf{42.43} & \textbf{2.07} \\
\bottomrule
\end{tabular}
\end{adjustbox}

\caption{Ablation study of HyLoVQA on VQA v2 and NExT-QA. Align. indicates adding the alignment loss term \( \mathcal{L}_{align} \) to the training objective.}
\label{tab:ablation}
\end{table}

\begin{figure}[t]
\centering
\includegraphics[width=1.0\linewidth]{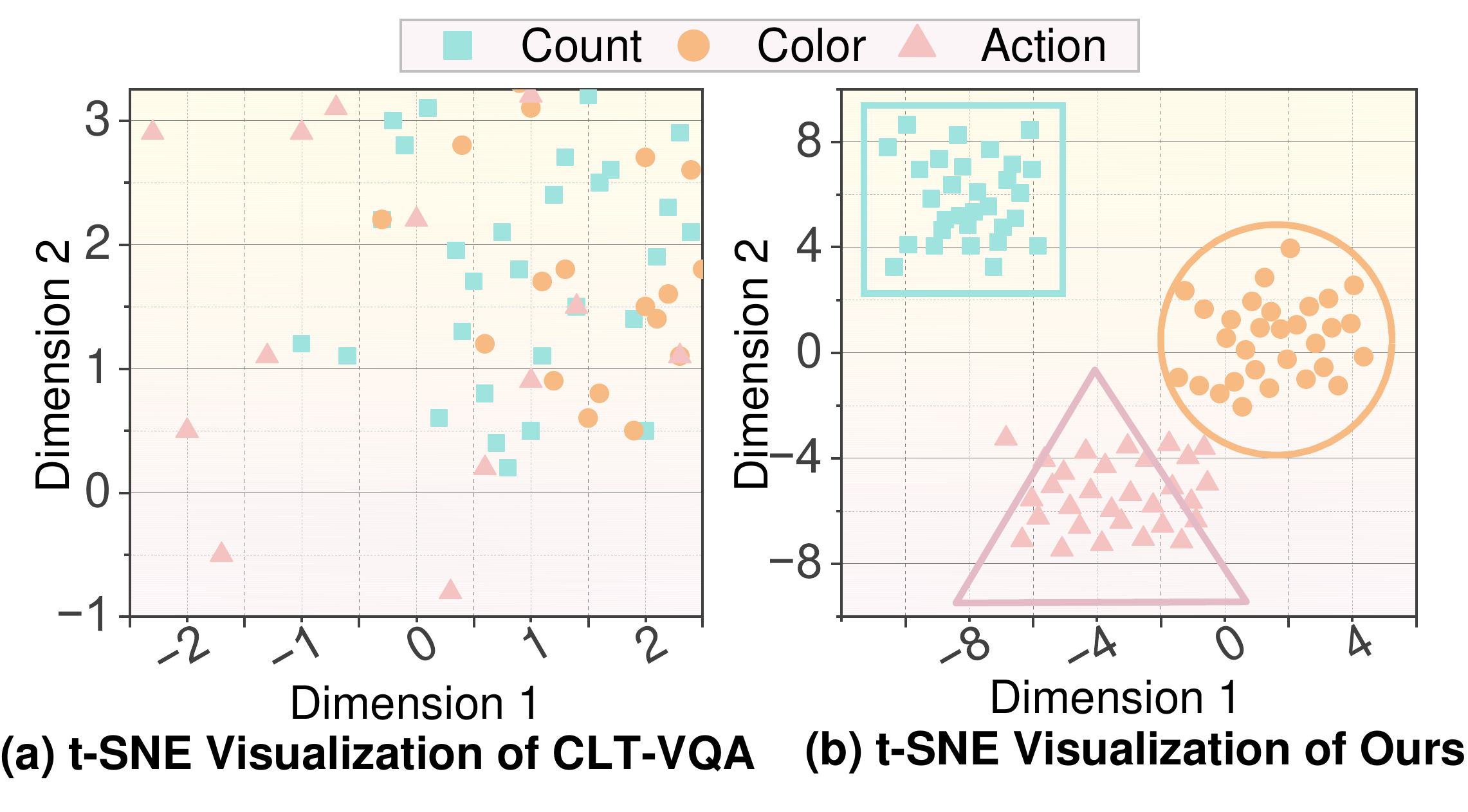} 
\caption{t-SNE visualization of sample feature representations on VQA v2 across task categories.}
\label{fig:tsne}
\end{figure}

\subsection{Experimental Settings}
\label{sec:exp_setup}

\noindent\textbf{Datasets.}
We evaluate on VQA v2 (200k+ COCO images and 1.1M QA pairs~\cite{35}) and NExT-QA, a video QA benchmark requiring temporal and causal reasoning~\cite{36}. Following the standard continual VQA protocol, VQA v2 is split into 10 question-type tasks (\textit{recognition, location, judge, commonsense, count, action, color, type, subcategory, causal}), and NExT-QA into 8 tasks (\textit{CW, TN, TC, DL, DB, DC, DO, CH}).

\noindent\textbf{Evaluation Metrics.} We report Final Average Performance (AP) and Average Forgetting (AF), which measure overall performance after continual learning and forgetting on earlier tasks, respectively\cite{55}.

\noindent\textbf{Baselines.} We compare HyLoVQA with sequential fine-tuning (Vanilla), EWC\cite{EWC}, MAS\cite{MAS}, ER\cite{ER}, DER\cite{DER}, VS\cite{VS}, VQACL\cite{VQACL}, QUAD\cite{QUAD}, ProtoGroup\cite{ProtoGroup} and CLT-VQA\cite{CLT-VQA}.
All methods are evaluated under the same task stream, backbone, and metrics for fair comparison.

\noindent\textbf{Implementation Details.}
Evaluation follows two paradigms: \emph{Standard} testing on seen task--object combinations and \emph{Novel Composition} testing on held-out (unseen) combinations. For visual tokens, we use a Faster R-CNN\cite{56} pretrained on Visual Genome to extract $N{=}36$ RoI features per image on VQA v2, and an inflated 3D ResNeXt-101\cite{57} to extract $N{=}16$ clip-level motion features on NExT-QA. We use Adam~\cite{53} with a learning rate of 3e$-$5, gradient clipping of 5, and a warmup ratio of 0.1; $\gamma_{1}$ and $\gamma_{2}$ are tuned from \{0.1, 1, 10\}. We implement our proposed method based on PyTorch.




\begin{figure}[!t]
\centering
\includegraphics[width=0.95\columnwidth]{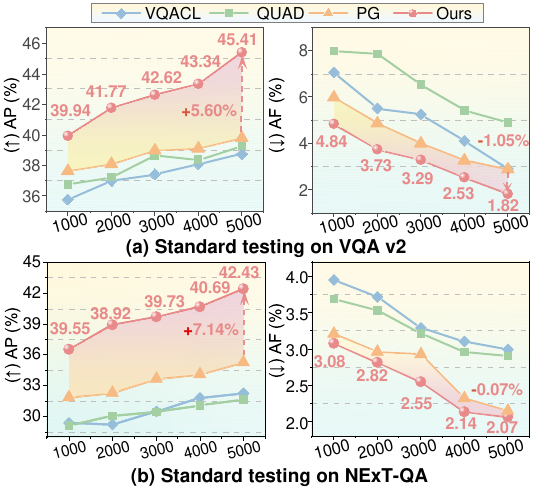} 
\caption{Memory size sensitivity analysis under standard testing on two datasets: (a) VQA v2 and (b) NExT-QA.}
\label{fig:mem_sensitivity}
\end{figure}

\subsection{RQ1: Main Results}
\label{sec:rq1}

As shown in Table~\ref{tab:main_results}, HyLoVQA achieves the best overall performance on both datasets under Standard and Novel Composition evaluation. On VQA v2, HyLoVQA attains 45.41\% AP/2.82\% AF (Standard) and 44.96\% AP/2.62\% AF (Novel), yielding about +4.72\% AP and +4.96\% AP over the strongest baseline, respectively, while maintaining low forgetting. On NExT-QA, HyLoVQA reaches 42.43\% AP/2.07\% AF (Standard) and 43.05\% AP/3.10\% AF (Novel), improving AP by about +6.66\% and +7.19\% over the best baseline in the two settings. Table~\ref{tab:group_results} further breaks down the Novel Composition setting under the group-removed protocol in Fig.\ref{fig:overview} (AP only), where we hold out one object group $G_i$ during training and evaluate Novel compositions involving $G_i$ versus Seen compositions over the remaining groups. HyLoVQA achieves the best AP on both Novel and Seen splits (VQA v2: 44.96\%/45.30\%; NExT-QA: 43.05\%/43.97\%), confirming consistent gains across group splits and stronger compositional generalization.

\subsection{RQ2: Ablation Study}
\label{sec:rq2}
Table~\ref{tab:ablation} shows that each component consistently contributes to continual VQA under the Standard Test setting. The values in parentheses denote the change relative to the full model (DMA+HyGen+Align.).
Starting from DMA only, performance drops and forgetting increases (41.77\% AP/3.46\% AF on VQA v2; 39.14\%/3.79\% on NExT-QA). Adding HyGen improves both AP and AF (44.96\%/2.95\% on VQA v2; 41.65\%/2.52\% on NExT-QA), confirming that anchor-conditioned, hypernetwork-generated LoRA adapters enhance instance-specific adaptation and alleviate interference. Adding Align. further yields the best trade-off (45.41\%/2.82\% on VQA v2; 42.43\%/2.07\% on NExT-QA) by keeping LoRA adapters aligned with the current task and object and reducing off-target adaptation.

\subsection{RQ3: Representation Stability (t-SNE)}
\label{sec:rq3}

This section examines whether HyLoVQA can efficiently adapt to the current sample input under continual distribution shift, where each sample consists of a visual object and a textual task.
As shown in Figure~\ref{fig:tsne}, we visualize sample features on VQA v2 and color points by task category (\emph{Count/Color/Action}).
Baseline features exhibit substantial overlap across categories, while HyLoVQA yields more compact intra-category clusters and clearer inter-category separation.
This indicates reduced cross-task representation entanglement, which is closely associated with forgetting and negative transfer in continual VQA.
We attribute this behavior to the DMA providing stable anchor references and HyGen performing anchor-conditioned adaptation, consistent with the higher AP and lower AF in Table~\ref{tab:main_results}.

\begin{figure}[!t]
\centering
\includegraphics[width=1.0\columnwidth]{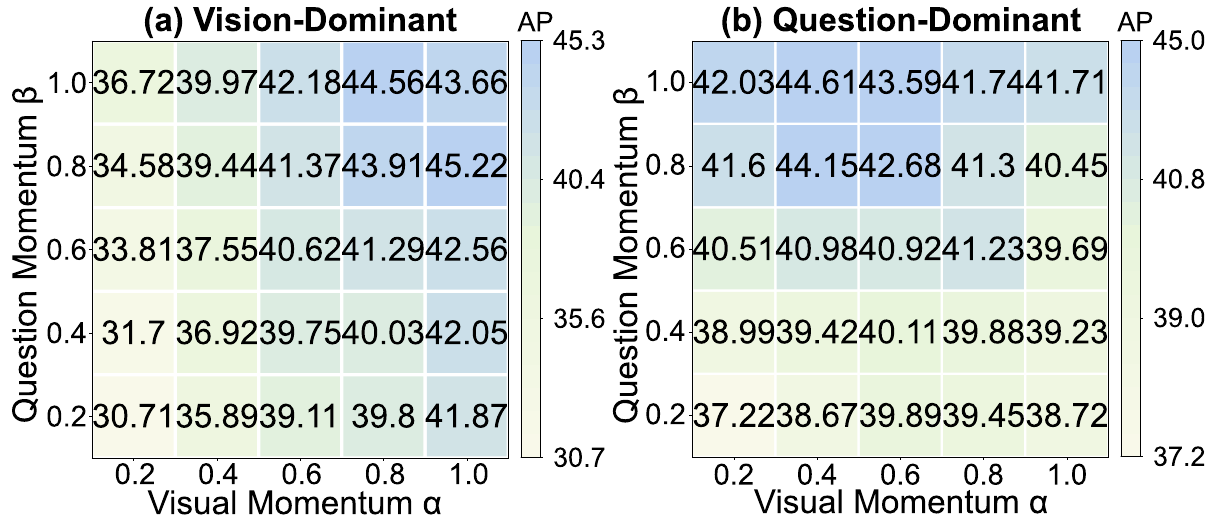} 
\caption{Sensitivity to modality-aware anchor momentum.}
\label{fig:heatmap_momentum}
\end{figure}

\subsection{RQ4: Memory Capacity Robustness}
\label{sec:rq4}

To evaluate memory capacity robustness, we vary the anchor memory bank size from 1k to 5k and report AP/AF under Standard testing on two datasets: VQA v2 and NExT-QA (Figure~\ref{fig:mem_sensitivity}).
HyLoVQA consistently outperforms strong baselines (VQACL, QUAD, and ProtoGroup) across all capacities, achieving a better AP--AF trade-off without requiring a large replay buffer.
On VQA v2 (Figure~\ref{fig:mem_sensitivity}a), increasing the memory from 1k to 5k improves AP from 39.94\% to 45.41\% while reducing AF from 4.84\% to 1.82\%.
On NExT-QA (Figure~\ref{fig:mem_sensitivity}b), AP increases from 39.55\% to 42.43\% and AF decreases from 3.08\% to 2.07\% as the memory grows from 1k to 5k.
Overall, HyLoVQA benefits steadily from additional memory across datasets, indicating effective and robust utilization of the anchor memory bank.

\subsection{RQ5: Hyperparameter Sensitivity}
\label{sec:rq5}
We evaluate the sensitivity of the modality-aware anchor update momentum on NExT-QA.
Figure~\ref{fig:heatmap_momentum} reports AP (\%) over a grid of visual and question momentum $(\alpha,\beta)$ and shows clear task-specific preferences, highlighting the flexibility of our DMA design.
On Vision-Dominant tasks (Figure~\ref{fig:heatmap_momentum}(a)), performance concentrates in the high-$\alpha$ region and peaks at $(\alpha,\beta)=(1.0,0.8)$ with 45.22\% AP.
In contrast, Question-Dominant tasks (Figure~\ref{fig:heatmap_momentum}(b)) favor larger $\beta$, achieving the best AP of 44.61\% at $(\alpha,\beta)=(0.4,1.0)$.
Overall, the optimal region shifts consistently with the dominant modality, validating that DMA can adapt its anchor/prototype update dynamics to diverse reasoning needs via modality-aware momentum control.

\begin{figure}[t]
\centering
\includegraphics[width=0.95\linewidth]{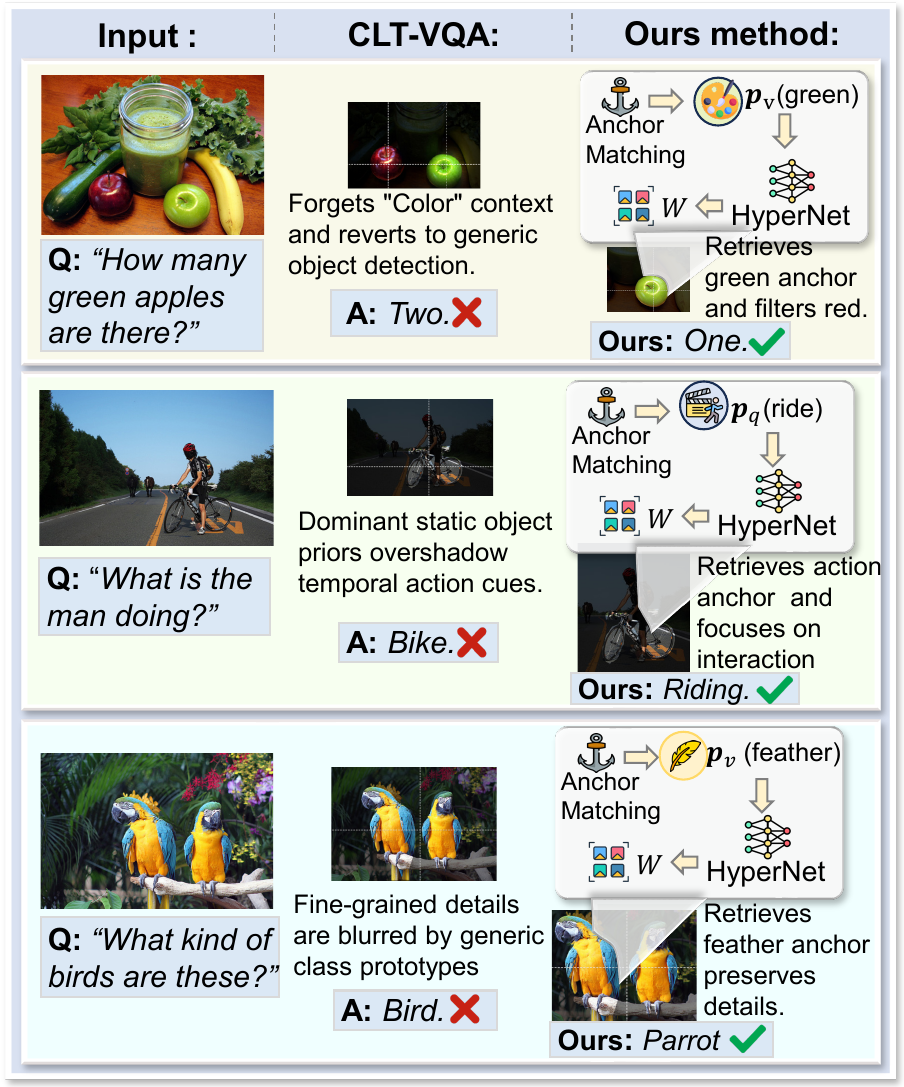} 
\caption{Qualitative case studies: CLT-VQA vs. HyLoVQA.}
\label{fig:case_study}
\end{figure}

\subsection{RQ6: Qualitative Case Study}
\label{sec:rq6}

Figure~\ref{fig:case_study} illustrates failure modes of prior methods and how HyLoVQA addresses them. CLT-VQA often forgets sample-specific constraints and falls back to generic shortcuts, e.g., ignoring color-conditioned counting, relying on static object priors in action queries, or merging fine-grained categories into coarse labels. HyLoVQA retrieves relevant anchors from DMA and uses HyGen to generate anchor-conditioned LoRA adapters that steer cross-attention toward task- and object-specific cues: retrieving $\boldsymbol{p}_v$ (\emph{green}) to suppress distractors, retrieving $\boldsymbol{p}_q$ (\emph{ride}) to emphasize interaction regions for action recognition, and retrieving fine-grained visual anchors to preserve discriminative feather details. As a result, HyLoVQA preserves the intended constraints under drift and produces answers consistent with the queried attributes and interactions (e.g., \emph{one}, \emph{riding}, and \emph{parrot}).

\section{Conclusion}
We propose HyLoVQA, a lightweight continual VQA framework that mitigates cross-level task interference while enabling dynamic adaptation to each incoming sample’s task and object. With a drift-resilient memory anchor bank and a hypernetwork that generates task/object-specific LoRA adapters, HyLoVQA supports dynamic and parameter-efficient adaptation and better preserves prior knowledge. A semantic--functional alignment loss further keeps updates focused on the current task and object. Overall, HyLoVQA highlights memory-guided parameter generation as a simple and general route to balance stability and plasticity in continual VQA under non-stationary and compositional shifts.

\section*{Acknowledgments}
This work was supported in part by the Natural Science Foundation of Hubei Province of China (No. 2025AFB653), the Open Fund of Hubei Key Laboratory of Big Data Intelligent Analysis and Application, Hubei University (No. 2025BDIAA01), and the National Natural Science Foundation of China (No. 62207011, 62407013, 62377009).

\bibliographystyle{named}
\bibliography{ijcai26}

\end{document}